\documentclass[11pt]{article}
\usepackage{acl2016}
\usepackage{times}
\usepackage{latexsym}
\usepackage{amsmath}
\usepackage{graphicx}

\aclfinalcopy % Uncomment this line for the final submission
\title{Improving Reliability of Word Similarity Evaluation\\ by Redesigning Annotation Task and Performance Measure}

\author{Oded Avraham \and Yoav Goldberg \\
		Computer Science Department \\
	    Bar-Ilan University\\       
	    Ramat-Gan, Israel\\
	    {\tt \{oavraham1,yoav.goldberg\}@gmail.com}}
\date{}

\begin{document}

\maketitle

\begin{abstract}
    %We survey some deficiencies in current word-similarity evaluation datasets,
    %and suggest an new method for constructing word-similarity datasets
    %that does not suffer from these deficiencies. 
We suggest a new method for creating and using gold-standard datasets for word similarity evaluation. Our goal is to improve the reliability of the evaluation, and we do this by redesigning the annotation task to achieve higher inter-rater agreement, and by defining a performance measure which takes the reliability of each annotation decision in the dataset into account.

\end{abstract}

\section{Introduction}

Computing similarity between words is a fundamental challenge in natural
language processing. Given a pair of words, a similarity model $sim(w_1,w_2)$
should assign a score that reflects the level of similarity between them, e.g.:
$sim(\textit{singer}, \textit{musician}) = 0.83$.  While many methods for computing $sim$
exist (e.g., taking the cosine between vector embeddings derived by word2vec
\cite{mikolov2013efficient}), there are currently no reliable measures of quality for such models.
In the past few years, word similarity models show a
consistent improvement in performance when evaluated using the conventional evaluation methods and datasets.
But are these evaluation measures really reliable indicators of the model
quality? Lately, Hill et al ~\shortcite{hill2015simlex} claimed that the answer
is no. They identified several problems with the existing datasets, and created
a new dataset -- SimLex-999 -- which does not suffer from them. However, we
argue that there are inherent problems with conventional datasets and the method
of using them that were not addressed in SimLex-999. We list these problems, and suggest a new and more reliable way of
evaluating similarity models. 
We then report initial experiments on a dataset of Hebrew nouns similarity that
we created according to our proposed method.
%Following these principles, we created a dataset of Hebrew words similarity.

\section{Existing Methods and Datasets for Word Similarity Evaluation}

Over the years, several datasets have been used for evaluating word similarity
models.  Popular ones include RG~\cite{rubenstein1965contextual}, WordSim-353~\cite{finkelstein2001placing}, WS-Sim~\cite{agirre2009study} and MEN~\cite{bruni2012distributional}.
Each of these datasets is a collection of word pairs together with their similarity scores as assigned by human annotators.
A model is evaluated by assigning a similarity score to each pair, sorting the
pairs according to their similarity, and calculating
the correlation (Spearman's $\rho$) with the human ranking. 

Hill et al~\shortcite{hill2015simlex} had made a comprehensive review of these datasets,
and pointed out some common shortcomings they have.
The main shortcoming discussed by Hill et al is the handling of
\textit{associated but dissimilar} words, e.g. (\textit{singer, microphone}): in datasets which contain such pairs (WordSim and MEN) they are
usually ranked high, sometimes even above pairs of similar words. This causes an
undesirable penalization of models that apply the correct behavior (i.e., always
prefer similar pairs over associated dissimilar ones). Other datasets (WS-Sim and
RG) do not contain pairs of associated words pairs at all. Their absence makes
these datasets unable to evaluate the models' ability to distinct between
associated and similar words. Another shortcoming  mentioned by Hill et al
~\shortcite{hill2015simlex} is \textit{low inter-rater agreement} over the human
assigned similarity scores, which might have been caused by unclear instructions for the annotation task.
As a result, state-of-the-art models reach the agreement ceiling for most of the datasets, while a simple manual evaluation will suggest that these models are still inferior to humans. In order to solve these shortcomings, Hill et al \shortcite{hill2015simlex}
developed a new dataset -- Simlex-999 -- in which the instructions presented to the annotators emphasized the difference
between the terms associated and similar, and managed to solve the discussed problems. 

While SimLex-999 was definitely a step in the right direction, we argue that
there are more fundamental problems which all conventional methods, including
SimLex-999, suffer from. In what follows, we describe each one of these problems.

\section{Problems with the Existing Datasets}
Before diving in, we define some terms we are about to use. Hill et al
~\shortcite{hill2015simlex} used the terms \textit{similar} and \textit{associated but dissimilar}, which they didn't formally connected to fine-grained semantic relations. However, by inspecting the average score per relation, they found a clear preference for hyponym-hypernym pairs (e.g. the scores of the pairs (\textit{cat, pet}) and (\textit{winter, season}) are much higher than those of the cohyponyms pair (\textit{cat, dog}) and the antonyms pair (\textit{winter, summer})). Referring hyponym-hypernym pairs as \textit{similar} may imply that a good
similarity model should prefer hyponym-hypernym pairs over pairs of other relations, which is not
always true since the desirable behavior is task-dependent. Therefore, we will
use a different terminology: we use the term 
 \textit{preferred-relation} to denote the relation which the model should
 prefer, and \textit{unpreferred-relation} to denote any other relation.
 
 The first problem is \textit{the use of rating scales}. Since the level of
 similarity is a relative measure, we would expect the annotation task to ask
 the annotator for a ranking. But in most of the existing datasets, the
 annotators were asked to assign a numeric score to each pair (e.g. 0-7 in
 SimLex-999), and a ranking was derived based on these scores. This choice is
 probably due to the fact that a ranking of hundreds of pairs is an exhausting
 task for humans. However, using rating scales makes the annotations vulnerable
 to a variety of biases \cite{friedman1999rating}. Bruni et al
 \shortcite{bruni2012distributional} addressed this problem by asking the
 annotators to rank each pair in comparison to 50 randomly selected pairs. This
 is a reasonable compromise, but it still results in a daunting annotation task,
 and makes the quality of the dataset depend on a random selection of comparisons.
 
The second problem is \textit{rating different relations on the same scale}. In Simlex-999, the annotators were instructed to assign low scores to unpreferred-relation pairs, but the decision of \textit{how low} was still up to the annotator.
While some of these pairs were assigned very low scores (e.g.
sim(\textit{smart}, \textit{dumb}) = 0.55), others got significantly higher ones (e.g. sim(\textit{winter},
\textit{summer}) = 2.38). A difference of 1.8 similarity scores should not be
underestimated --  in other cases it testifies to a true superiority of one
pair over another, e.g.: sim(\textit{cab}, \textit{taxi}) = 9.2, sim(\textit{cab}, \textit{car}) = 7.42. The
situation where an arbitrary decision of the annotators 
affects the model score, impairs the reliability of the evaluation: a model shouldn't be punished for preferring (\textit{smart, dumb}) over (\textit{winter, summer}) or vice versa, since this comparison is just ill-defined.

The third problem is \textit{rating different target-words on the same scale}.
Even within preferred-relation pairs, there are ill-defined comparisons, e.g.:
(\textit{cat, pet}) vs. (\textit{winter, season}). It's quite unnatural to
compare between pairs that have different target-words, in contrast to pairs which share the target word, like (\textit{cat, pet}) vs. \textit{cat, animal}).
Penalizing a model for preferring (\textit{cat, pet}) over (\textit{winter, season}) or vice versa impairs the evaluation reliability.

The fourth problem is that \textit{the evaluation measure does not consider annotation decisions reliability}.
The conventional method measures the model score by calculating Spearman
correlation between the model ranking and the annotators average ranking. This
method ignores an important information source: the reliability of each
annotation decision, which can be determined by the agreement of the annotators
on this decision. For example, consider a dataset containing the pairs (\textit{singer, person}), (\textit{singer, performer}) and (\textit{singer, musician}). Now let's assume that in
the average annotator ranking, (\textit{singer, performer}) is ranked above (\textit{singer, person}) after 90\% of the annotators assigned it with a higher score, and
(\textit{singer, musician}) is ranked above (\textit{singer, performer}) after 51\% percent of the
annotators assigned it with a higher score. Considering this, we would like the
evaluation measure to severely punish a model which prefers (\textit{singer, person})
over (\textit{singer, performer}), but be almost indifferent to the model's decision over
(\textit{singer, performer}) vs. (\textit{singer, musician}) because it seems that even humans
cannot reliably tell which one is more similar. In the conventional datasets, no
information on reliability of ratings is supplied except for the overall
agreement, and each average rank has the same weight in the evaluation measure.
The problem of reliability is addressed by Luong et al
\shortcite{luong2013better} which included many rare words in their dataset, and
thus allowed an annotator to indicate ``Don't know'' for a pair if they does not know one of the words. The problem with applying this approach as a more general reliability indicator is that the annotator confidence level is subjective and not absolute.

\section{Proposed Improvements}
We suggest the following four improvements for handling these problems.

\noindent (1) The annotation task will be an explicit ranking task. Similarly to
Bruni et al \shortcite{bruni2012distributional}, each pair will be directly
compared with a subset of the other pairs. Unlike Bruni et al, each pair will be
compared with only a few carefully selected pairs, following the
principles in (2) and (3).

\noindent (2) A dataset will be focused on a single preferred-relation type
(we can create other datasets for tasks in which the preferred-relation is
different), and only preferred-relation pairs will be presented to the annotators.
We suggest to spare the annotators the effort of considering the \textit{type}
of the similarity between words, in order to let them concentrate on the
\textit{strength} of the similarity.
Word pairs following unpreferred-relations will not be included in the
annotation task but will still be a part of the dataset -- we always add them to
the bottom of the ranking. For example, an annotator will be asked to rate (\textit{cab, car}) and (\textit{cab, taxi}), but not (\textit{cab, driver}) -- which will be ranked last since it's an unpreferred-relation pair.

\noindent (3) Any pair will be compared only with pairs sharing the same target word. We suggest to make the pairs ranking more reliable by splitting it into multiple target-based rankings, e.g.: (\textit{cat, pet}) will be compared with (\textit{cat, animal}), but not with (\textit{winter, season}) which belongs to another ranking.

\noindent (4) The dataset will include a reliability indicator for each
annotators decision, based on the agreement between annotators. The reliability
indicator will be used in the evaluation measure: a model will be penalized more
for making wrong predictions on reliable rankings than on unreliable ones.

\subsection{A Concrete Dataset}
In this section we describe the structure of a dataset which applies the above
improvements. First, we need to define the preferred-relation, in order to apply improvement (2).
In what follows we use the hyponym-hypernym relation as the preferred-relation.

The dataset is based on \textit{target words}.
For each target word we create a group of \textit{candidate words}, which we
refer to as the \textit{target-group}. Each candidate word belongs to one of
three categories: \textit{positives} (related to the target, and the type of the
relation is the preferred one), \textit{distractors} (related to the target, but
the type of the relation is not the preferred one), and \textit{randoms} (not
related to the target at all). For example, for the target word \textit{singer}, the target group may include
\textit{musician}, \textit{performer}, \textit{person} and \textit{artist} as
positives, \textit{dancer} and \textit{song} as distractors, and \textit{laptop}
as random. For each target word, the human annotators will be asked to rank the
positive candidates by their similarity to the target word (improvements (1) \& (3)).
For example, a possible ranking may be: \textit{musician} $>$ \textit{performer} $>$ \textit{artist} $>$ \textit{person}.

The annotators responses allow us to create the actual dataset, which consists
of a collection of \textit{binary comparisons}.
A binary comparison is a value $R_{>}(w_1,w_2 ; w_t)$ indicating how likely it
is to rank the pair ($w_t, w_1$) higher than ($w_t, w_2$), where $w_t$ is a
target word and $w_1$, $w_2$ are two candidate words.
By definition, $R_{>}(w_{1},w_{2};w_{t})$ = 1 - $R_{>}(w_{2},w_{1};w_{t})$.
For each target-group, the dataset will contain a binary comparison for any
possible combination of two positive candidates, as well
as for all the combinations in which the first candidate is positive and the second is negative (either distractor or
random). When comparing two positive candidates $w_{p_1}$,$w_{p_2}$ --  the value of
$R_{>}(w_{p_1},w_{p_2};w_{t})$ is the portion of annotators who ranked ($w_{t}, w_{p_1})$ over ($w_{t}, w_{p_2}$). When comparing a positive candidate $w_{p}$ to a negative one $w_{n}$ -- the value 
of $R_{>}(w_{p},w_{n};w_{t})$ is 1.
%For each target-group, the dataset will contain a binary comparison for any
%possible combination of two positive complements $w_{p1}$ and $w_{p2}$, where the value of
%$R_{>}(w_{1},w_{2};w_{t})$ is the portion of annotators which ranked $(w_{t},
%w_{1})$ over $(w_{t},w_{2})$. It will also contain any possible combination of a
%positive complement with a negative one (distractor or random). In these
%comparisons, $R_{>}(w_{p},w_{n};w_{t})$ will be always assigned with 1 (where
%$w_{p}$ is a positive complement, and $w_{n}$ is a negative one).
This reflects the intuition that a good model should always rank preferred-relation pairs above other pairs. 
Notice that $R_{>}(w_{1},w_{2};w_{t})$ is the reliability indicator for each of the dataset key answers, which will be used to apply improvement (4).
For some example comparisons, see Table \ref{dataset-table}.

\begin{table}
\small
\centering
\begin{tabular}{l|l|l|l|c}
\hline
& $w_{t}$ & $w_{1}$ & $w_{2}$ & $R_{>}(w_{1},w_{2};w_{t})$ \\\hline
P & singer & person & musician & 0.1 \\
P & singer & artist & person & 0.8 \\
P & singer & musician & performer & 0.6 \\
D & singer & musician & song & 1.0 \\
R & singer & musician & laptop & 1.0 \\\hline
\end{tabular}
\caption{\label{dataset-table}\footnotesize Binary Comparisons for the target word
\textit{singer}. P: positive
pair; D: distractor pair; R: random pair.}
\end{table}

\subsection{Scoring Function}
Given a similarity function between words $sim(x,y)$ and a triplet $(w_t, w_1,
w_2)$ let $\delta=1$ if $sim(w_t, w_1) > sim(w_t, w_2)$ and $\delta=-1$
otherwise. The score $s(w_t, w_1, w_2)$ of the triplet is then:
$s(w_t, w_1, w_2) = \delta(2R_{>}(w_1,w_2;w_t)-1)$.
This score ranges between $-1$ and $1$, is positive if the model ranking agrees
with more than 50\% of the annotators, and is $1$ if it agrees with all of them.
The score of the entire dataset $C$ is then:
\[
\frac{\sum_{w_t,w_1,w_2\in C}\max(s(w_t,w_1,w_2),0)}{\sum_{w_t,w_1,w_2\in
C}|s(w_t,w_1,w_2)|}
\]

The model score will be 0 if it makes the wrong decision (i.e.
assign a higher score to $w_{1}$ while the majority of the annotators ranked
$w_{2}$ higher, or vice versa) in every comparison. If it always makes the right
decision, its score will be 1. Notice that the \textit{size} of the majority
also plays a role. When the model takes the wrong decision in a comparison,
nothing is being added to the numerator. When it takes the right decision, the
numerator increase will be larger as reliable as the key answer is,
and so is the general score (the denominator does not depend on the model
decisions). 

It worth mentioning that a score can also be computed over a subset of $C$, as comparisons of specific type (positive-positive,
positive-distractor, positive-random). This allows the user of the dataset to
make a finer-grained analysis of the evaluation results: it can get the quality of the
model in specific tasks (preferring similar words over less similar, over words from unpreferred-relation, and over random words) rather than just the general quality.

\section{Experiments}
We created two datasets following the proposal discussed above:
one preferring the hyponym-hypernym relation, and the other the cohyponym
relation.\footnote{Our datasets and evaluation script are available on \textit{https://github.com/oavraham1/ag-evaluation}}
The datasets contain Hebrew nouns, but such datasets can be created for
different languages and parts of speech -- providing that the language has basic lexical resources.
For our dataset, we used a dictionary, an encyclopedia and a thesaurus to create
the hyponym-hypernym pairs, and databases of word association norms \cite{rubinsten2005free} and categories norms \cite{henik1988category} to create the distractors pairs and the cohyponyms pairs, respectively. The hyponym-hypernym dataset is based on 75
target-groups, each contains 3-6 positive pairs, 2 distractor pairs and one
random pair, which sums up to 476 pairs. The cohyponym dataset is based on 30
target-groups, each contains 4 positive pairs, 1-2 distractor pairs and one random
pair, which sums up to 207 pairs. 

\begin{figure*}
\centering
\includegraphics[width=0.9\textwidth]{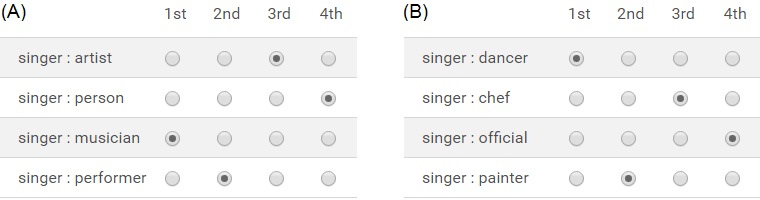}
\caption{\label{fig:example}The example rankings we supplied to the annotators as a part of the questionnaires instructions (translated from Hebrew). Example (A) appeared in the hyponym-hypernym questionnaires, while (B) appeared in the cohyponyms  questionnaire.}
\end{figure*}

We used the target groups to create 4 questionnaires: 3 for the hyponym-hypernym relation (each contains 25 target-groups), and one for the cohyponyms relation. We asked human annotators to order the positive pairs of each target-group by the
similarity between their words. In order to prevent the annotators from confusing between the different aspects of similarity, each annotator was requested to answer only one of the questionnaires, and the instructions for each questionnaire included an example question which demonstrates what the term ``similarity'' means in that questionnaire (as shown in Figure \ref{fig:example}).

Each target-group was ranked by 18-20
annotators. We measured the average pairwise inter-rater agreement, and as done
in \cite{hill2015simlex} -- we excluded any annotator which its agreement with
the other was more than one standard deviation below that average (17.8 percent
of the annotators were excluded). The agreement was quite high (0.646 and 0.659 for hyponym-hypernym and cohyponyms target-groups, respectively), especially considering that in contrast to other datasets -- our annotation task did not include pairs that are ``trivial'' to rank (e.g. random pairs). Finally, we used the remaining annotators
responses to create the binary comparisons
collection. The hyponym-hypernym dataset includes 1063 comparisons, while the
cohyponym dataset includes 538 comparisons. 

To measure the gap between a human and a model performance on the dataset, we trained a word2vec \cite{mikolov2013efficient} model \footnote{We used code.google.com/p/word2vec implementation, with window size of 2 and dimensionality of 200.} on the Hebrew Wikipedia. We used two methods of measuring: the first is the conventional way (Spearman correlation), and the second is the scoring method we described in the previous section, which we used to measure general and per-comparison-type scores. The results are presented in Table \ref{score-table}.

\begin{table}
\small
\centering
\begin{tabular}{|l|c|c|}
\hline
 & Hyp. & Cohyp. \\\hline
Inter-rater agreement & 0.646 & 0.659 \\\hline
w2v correlation & 0.451 & 0.587 \\\hline
w2v score (all) & 0.718 & 0.864 \\
w2v score (positive) & 0.763 & 0.822 \\
w2v score (distractor) & 0.625 & 0.833 \\
w2v score (random) & 0.864 & 0.967 \\\hline
\end{tabular}
\caption{\label{score-table} \footnotesize The hyponym-hypernym dataset agreement (0.646) compares favorably with the
agreement for nouns pairs reported by Hill et al \shortcite{hill2015simlex} (0.612), and it is much higher than the correlation score of the word2vec model. Notice that useful insights can be gained from the
per-comparison-type analysis, like the model's difficulty to distinguish hyponym-hypernym pairs from other relations.}
\end{table}

\section{Conclusions}
We presented a new method for creating and using datasets for word similarity, which improves evaluation reliability by redesigning the annotation task and the performance measure. We created two datasets for Hebrew and showed a high inter-rater agreement.
Finally, we showed that the dataset can be used for a finer-grained analysis of the model quality. A future work can be applying this method to other languages and relation types.

\section*{Acknowledgements}
The work was supported by the Israeli Science Foundation (grant number 1555/15). We thank Omer Levy for useful discussions.

\bibliography{acl2016}
\bibliographystyle{acl2016}

\end{document}